# Mitigating Presentation Attack using DCGAN and Deep CNN

Nyle Siddiqui and Rushit Dave

Dept. of Computer Science, University of Wisconsin Eau Claire, WI, USA.

**Abstract**

*Biometric based authentication is currently playing an essential role over conventional authentication system; however, the risk of presentation attacks subsequently rising. Our research aims at identifying the areas where presentation attack can be prevented even though adequate biometric image samples of users are limited. Our work focusses on generating photorealistic synthetic images from the real image sets by implementing Deep Convolution Generative Adversarial Net (DCGAN). We have implemented the temporal and spatial augmentation during the fake image generation. Our work detects the presentation attacks on facial and iris images using our deep CNN, inspired by VGGNet [1]. We applied the deep neural net techniques on three different biometric image datasets, namely MICHE-I [2], VISOB [3], and UBI-Pr [4]. The datasets, used in this research, contain images that are captured both in controlled and uncontrolled environment along with different resolutions and sizes. We obtained the best test accuracy of 97% on UBI-Pr [4] Iris datasets. For MICHE-I [2] and VISOB [3] datasets, we achieved the test accuracies of 95% and 96% respectively.*

## I. Introduction

Biometric-based presentation attacks are primarily focused on gaining access to a biometric sample from secured databases or external resources and illegally implementing the same biometric to gain access to secured data. Though biometric scanning strengthens security through unique features, copying the biometric sample and those unique features to illegally gain access to a biometric system is feasible. Presentation attack detection and mitigation is a challenge[12-19]. There are many ways to spoof a biometric system. High-resolution copies of biometric samples have been used as print attacks [5]. Synthetic images and digitally retouched images have also worked in presentation attacks [6]. Deep learning techniques in combination with Deep Convolution Generative Adversarial Net (DCGAN) [7] and modified VGG-Net [8] can be a good solution to mitigate the attack.

Research has been done to prevent iris attacks in large scale. The overall purpose of the study was to determine how vulnerable different computers are to handle the fake iris image-set [9]. Another study on DCGAN related to unlabeled images was conducted by a research group at the University of Technology Sydney. The purpose of their study was to determine if the unlabeled images generated through DCGAN would be able to authenticate a real person [1].

Recently deep learning approaches have been used to detect presentation attacks [8]. However, deep learning-based Presentation Attack Detection (PAD) methods suffer from over-fitting. To address this issue, this research uses DCGAN to create a realistic image dataset[21]. Deep neural network will then be used to detect attacks. In this research, we have implemented the temporal and spatial augmentation during the fake image generation.

Deep convolutional generative adversarial network (DCGAN) is a convolutional neural network that works primarily under unsupervised environment, and generate photorealistic synthetic images from the real images, where the real images are not adequate. This Network can be used for full images strictly focused on identifiable features in it. Specifically, for this research our goal is to extract synthetic photorealistic images from the available real images to train the deep convolution net. The architecture of the DCGAN are unique compared to all the other CNNs as it deals with the noise generator during the process of generating synthetic fake images. The spatial and temporal augmentation also plays a pivotal role here during the fake image generation.

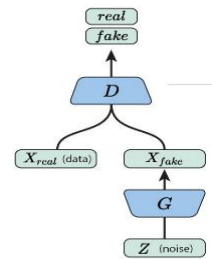

Fig1: Generative Adversarial Networks (GANs) illustration.

We select our input image datasets under controlled and uncontrolled environments, for the real time training, validation, and testing. Our already proposed modified VGGNet [8,2022], has a total of 14 discrete Convolution 2D layers, 5 Max Pooling 2D Layers, 1 Flatten layer, and 2 Dense2D layers. In this research, we use sigmoid function instead of soft-max approach for the faster 2D binary classification and validation. In Fig 2, we briefly explained the main components of our 'modified VGGNet' structure. Also, the CNN architecture used in this paper is shown in Fig. 3.

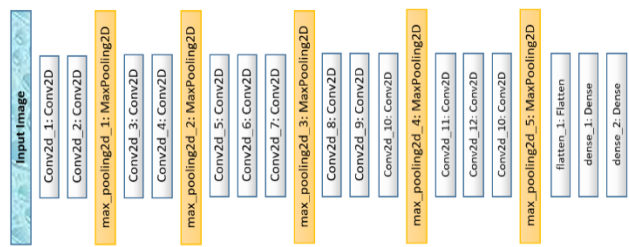

Fig 2 Internal Structure of Modified VGGNet [8]

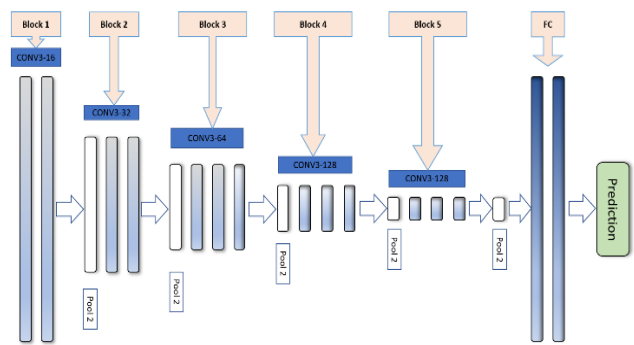

Fig 3 Architecture of Modified VGGNet [8]

### III. Proposed Approach

We have developed an architecture using a deep Convolutional Neural Network (CNN) for our research to prevent presentation attack. We use the datasets that classify the real iris images with the images synthetically generated by DCGAN. We apply our method on the MICHE-I [2, 32], VISOB [3], and UBI periocular [4] dataset which contain real images. We collected DCGAN generated synthetic images. We decide to collect two hundred real images in each iteration and generated around 10,000 synthetic images altogether. Interestingly, MICHE-I [2], VISOB [3] and UBI-Pr [4] dataset has equal ratio of left and right iris images. We have applied the same techniques on MICHE-I [2], VISOB [3] and UBI-Pr [4,7] datasets to generate the synthetic photorealistic image sets. We distribute the real images and synthetic images for each dataset separately and implement them during the training process in the modified VGG-Net [8] and validate them afterward. We split the training set and test set in 80 – 20 ratios overall.

### IV. Results and Discussion

The training pattern of CNNs varies depending on time and memory availability of CPU/GPUs. To achieve reasonable performances, we ran different implementations CNNs, including the modified VGGNet, on all the datasets separately for multiple times and observed their performances and minute variations. We made a split to the datasets and used 80% for training purposes and 20% for testing purposes. In Table 1, we compare the test accuracies of our 'modified VGGNet' to the conventional Alex-Net [10,31] and Spoof Net. The best results we achieved using our 'modified VGGNet' on different segment of image datasets are shown in the Table 1.

Our research reveals result on the performance and accuracies on MICHE-I [2], VISOB [3], and UBI-Pr [4] datasets with the significantly high accuracies. Though our modified VGG Net outperforms the other two classifiers, i.e. AlexNet [10] and Spoof Net [11] in all datasets (see Fig 4).

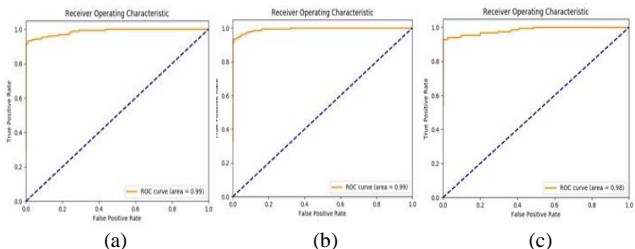

Fig 4: ROC Curve on our proposed modified VGGNets for the following datasets (a) MICHE-I (Acc: 95%), (b) UBI-Pr Dataset (Acc: 97%), (c) VISOB datasets (Acc: 96%).

| CNNs Architecture | Datasets | | |
|---|---|---|---|
| | MICHE-I | UBI-Pr | VISOB |
| Modified VGG Net | 95% | 97% | 96% |
| Alex Net | 93% | 91% | 90% |
| Spoof Net | 94% | 90% | 92.1% |

Table 1: Classification accuracies of different implementations of CNNs.

### V. Conclusion and Future Work

In conclusion we discover that the UBI periocular left and right eye images developed have higher accuracy than all the other datasets. We are also able to generate photorealistic synthetic image sets from each dataset using DCGAN [23-30] in large scale and implemented thereafter. Furthermore, that the modified VGGNet [8] produced high true positive rate (in the range of 98 – 99 %). Our future work will primarily focus on time optimization on the DCGAN generated fake image-sets with higher resolution and their fast implementation during classification.

### VI. Acknowledgement

This work is supported in part by the National Science Foundation under Grant HRD-1719488 and National Security Agency under Grant H98230-18-1-0097. Any opinions, findings, and conclusions or recommendations expressed in this work are those of the authors and do not necessarily reflect the views of the funding agencies.